\documentclass[journal]{IEEEtran} % use the `journal` option for ITherm conference style
\usepackage{dblfloatfix}    % To enable figures at the bottom of page
\IEEEoverridecommandlockouts
\usepackage{cite}
\usepackage{amsmath,amssymb,amsfonts}
\usepackage{algorithmic}
\usepackage{graphicx}
\usepackage{textcomp}
\usepackage{xcolor}
\usepackage{multirow}
\usepackage{array}
\usepackage{longtable}
\usepackage{booktabs}
\usepackage{subfigure} 
\def\BibTeX{{\rm B\kern-.05em{\sc i\kern-.025em b}\kern-.08em
    T\kern-.1667em\lower.7ex\hbox{E}\kern-.125emX}}
\begin{document}

\title{Smart Mobility Digital Twin for Automated Driving: Design and Proof-of-Concept \\
}

\author{%%%% author names
    \IEEEauthorblockN{Kui Wang}, \IEEEauthorblockN{Zongdian Li}, \IEEEauthorblockN{Tao Yu}, \IEEEauthorblockN{Kei Sakaguchi}
    
    \IEEEauthorblockA{Tokyo Institute of Technology, Tokyo, Japan}
    
    \IEEEauthorblockA{Email: wang.k.ag@m.titech.ac.jp, \{yutao, lizd, sakaguchi\}@mobile.ee.titech.ac.jp}}

\maketitle

\begin{abstract}
During the past decade, smart mobility and intelligent vehicles have attracted increasing attention, because they promise to create a highly efficient and safe transportation system in the future. Meanwhile, digital twin, as an emerging technology, will play an important role in automated driving and intelligent transportation systems. This technology is applied in this paper to design a platform for smart mobility, providing large-scale route planning services. Utilizing sensing technologies and cloud/edge computing, we build a digital twin system model that reflects the static and dynamic objects from the real world in real time. With the smart mobility platform, we realize traffic monitoring and route planning through cooperative environment perception to help automated vehicles circumvent jams. A proof-of-concept test with a real vehicle in real traffic is conducted to validate the functions and the delay performance of the proposed platform.

\end{abstract}


\section{Introduction}
Thanks to the rapid development of communication technology and substantial capacity improvement, we can envision a future where data, knowledge, and resources can be seamlessly shared among automated vehicles and intelligent transportation systems (ITS)\cite{b1}. This leads to research interest in building a new system or platform to manage and process such huge volumes of traffic, sensing, and control data. 
In particular, from the perspective of traffic efficiency and driving safety, introducing the concept of a digital twin with centralized system architecture and distributed edge processing capability becomes essential. 
A digital twin platform enables instant local information processing and exchange at the edges, and a periodic global information update and aggregation in the cloud. Thus, while making decisions, the accuracy and completeness of the information can be guaranteed.

The advancements in vehicle-to-everything (V2X) communication and sensing technologies provide real-time information acquisition capabilities to establish digital twin systems. Today's V2X incorporates four types of interfaces: vehicle-to-vehicle (V2V), vehicle-to-infrastructure (V2I), vehicle-to-pedestrian (V2P) and vehicle-to-network (V2N) \cite{b2}. Via V2X, not only perception information but also the driving intentions of vehicles can be shared to improve traffic efficiency. On the other hand, an integrated sensing system composed of cameras, radars, and light detection and rangings (LiDARs) makes environmental perception more reliable and precise. Roadside units (RSUs) equipped with such sensing systems can monitor and collect traffic information at each intersection.

Therefore, studies related to mobility digital twin systems came forth recently. Authors in \cite{b3} summarized digital twins and their applications in connected and automated vehicles. Preliminary works in \cite{b4,b5,b6,b7,b8,b9} discussed the digital twin at architectural and theoretical levels. The authors in \cite{b4,b5,b6} made a step forward by implementing digital twin systems reliant upon the collection of real-life and real-time traffic information to support cloud-based advanced driver-assistance systems (ADAS) functions, predict traffic speed and analyze highway driving safety, respectively. A holistic mobility digital twin framework was presented in \cite{b7}, which consisted of three building blocks: the human digital twin for driver-type classification, the vehicle digital twin for cloud-based advanced ADAS, and the traffic digital twin for traffic flow monitoring.
However, few studies explicitly explain how the cloud and edges in a mobility digital twin cooperate to improve traffic efficiency and driving safety. There are pros and cons to utilizing computing and communication resources in the cloud and edges. The former provides stronger data processing capability with higher latency, which is suitable to handle large-scale and computation-intensive tasks such as dynamic traffic signal timing and long-term path planning; the edge indicates the RSUs and onboard units (OBUs), has better real-time performance to address delay-sensitive tasks such as risk alerting and vehicle maneuvering. Considering the importance of cloud and edge cooperation, it is necessary to design a novel digital twin architecture that mostly leverages communication and computing capabilities for smart mobility. In addition, the implementation of the digital twin architecture is also critical, through which the performances can be demonstrated and evaluated.
%%%%%%%%%%%%%%%%%%%%%%%%02/13%%%%%%%%%%%%%%%%%%%%%%%%%%%%%%%%

To deal with this challenge, a novel real-time digital twin platform is proposed to realize automated driving with good safety and high commuting efficiency by distributing different functional modules (i.e., navigation, environment perception, path planning, and control) of the single vehicular operating system to the cloud and the edges. Specifically, we present a digital twin model that reflects static and dynamic information from the physical space to the digital space and simultaneously visualizes 3D models of all objects in the sensors’ fields of view. 

The contributions of this paper lie in:
\begin{itemize}
    \item[1)] A real-time digital twin model integrates and visualizes static objects (e.g., buildings, infrastructures, and roads) and dynamic objects of interest (e.g., traffic participants including vehicles and pedestrians);
    \item[2)] A smart mobility digital twin platform is proposed to serve connected automated vehicles functionally based on cooperative perception and V2X communications;
    \item[3)] We exploit the cloud/edge communication and computing capabilities by allocating different functions and services over the cloud and the edge planes (environment perception at RSU and OBU edges, cooperative perception and route planning in the cloud, and motion planning as well as motion control in the vehicle edges).
    \item[4)] We implement the proposed platform to monitor real traffic and provide services for automated vehicles, as well as to study and evaluate its performances from the standpoints of functionality and latency.
\end{itemize}

The remainder of the paper is organized as follows: Section II describes the application scenarios, and presents the digital twin system architecture of the smart mobility platform. Details about the proof-of-concept, including hardware and software installation, and evaluation results, are shown in Section III. Section IV draws our concluding remarks.

\section{System Architecture}

In this section, we describe the application scenarios and requirements and propose the system architecture for building the digital twin model and the digital twin platform to provide services for smart mobility. 

\subsection{Application Scenario}

\begin{figure}[t]
    \centerline{\includegraphics[scale=0.5]{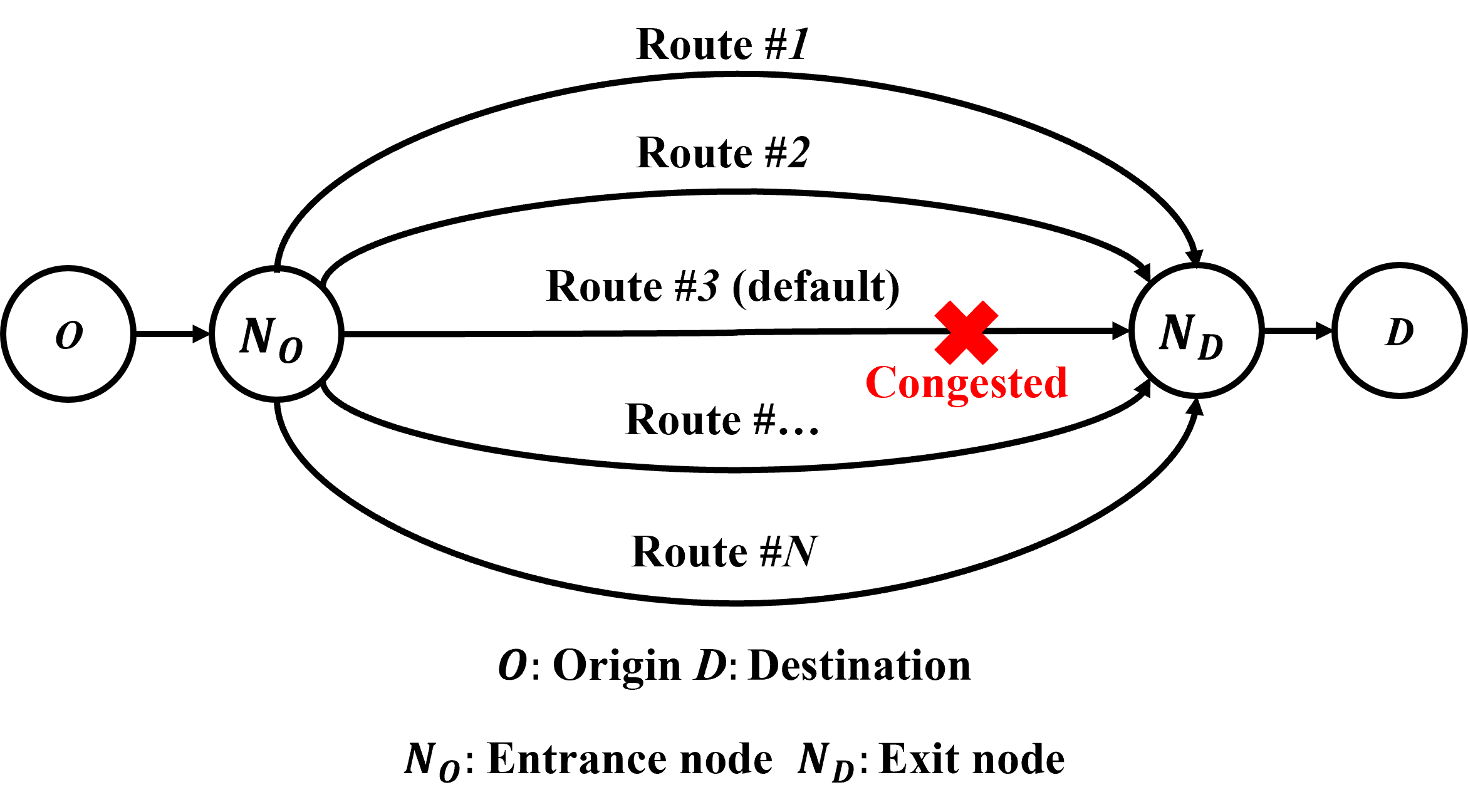}}
    \caption{Re-routing scenario.}
    \label{fig:routing-overview}
\end{figure}

The most important points in providing services to automated vehicles based on the proposed platform are: 1) achieving real-time traffic monitoring; and 2) long-term route planning based on traffic conditions. To better demonstrate these two points, a re-routing scenario is introduced and some specific requirements are discussed.

The re-routing scenario is often caused by traffic congestion. There are many reasons for traffic congestion, e.g., traffic accidents or excessive traffic and pedestrian flow. Considering both safety and efficiency, when traffic congestion occurs in the default route, it is necessary to avoid congested road sections by choosing another one. In the re-routing scenario, as shown in Fig.\ref{fig:routing-overview}, a vehicle moves from origin $O$ to destination $D$. The nearest traffic nodes/intersections from the origin and destination are named origin/entrance node $N_O$ and destination/exit node $N_D$, respectively. There are several routes for the ego vehicle from the origin node to the destination node, and the shortest route is regarded as the default one. When the degree of congestion, i.e., the road occupancy level, on this default route, is too high, a re-routing service will be triggered to choose another route with a lower road occupancy. In this scenario, the functions and key performance indicator (KPI) requirements of the digital twin platform can be summarized as follows:

\begin{itemize}
    \item[1)] Traffic environment perception and traffic flow monitoring on the RSU edges with low latency.
    \item[2)] The cloud integrates the traffic data and makes routing decisions.
    \item[3)] The routing decisions should be sent to the ego vehicle before it arrives at the origin node $N_{O}$. Assume that the ego vehicle drives with a constant speed $V_{ego}$ (m/s) and $S$ (m) away from the origin node. Then the total latency of the system $T_{total}$ should satisfy:
\begin{equation}
    T_{total} \leq \frac{S}{V_{ego}}
\end{equation}
\end{itemize}

\begin{figure}[t]
    \centerline{\includegraphics[scale=0.3]{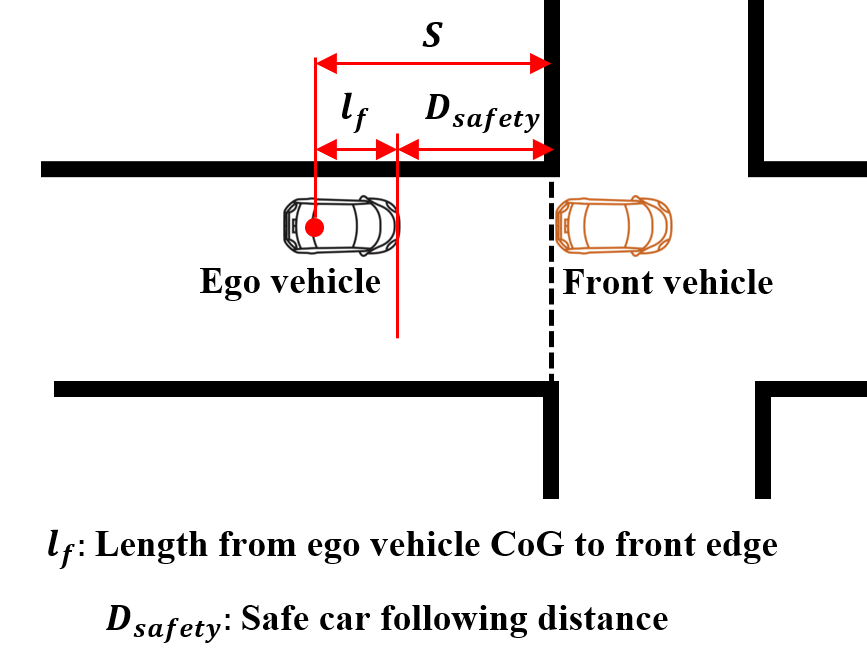}}
    \caption{Car following scenario.}
    \label{fig: situation}
\end{figure}

In order to minimize the impact of the future evolution in traffic, the trigger of route planning should meet the following two items:

\begin{itemize}
    \item[1)] The distance $S$ between the origin $O$ and the origin node $N_O$ should be as short as possible;
    \item[2)] In front of the ego vehicle, all other vehicles have entered the origin node, i.e., the ego vehicle should be the last one to enter the intersection area.
\end{itemize}

If there are no other vehicles in front of the ego vehicle, the planning decisions need only be sent to the ego vehicle before entering the intersection area. However, considering a car-following situation, the route planning service should be triggered at the moment the ego vehicle moves into the intersection as shown in Fig. \ref{fig: situation}. the distance from the origin to the intersection $S$ in equation (1) can be decided as:

\begin{equation}
    S = D_{safety} + l_r
\end{equation}

\noindent where $D_{safety}$ and $l_{r}$ are the safe following distance and the length from the center of gravity (CoG) of the ego vehicle to its front edge, respectively.

\subsection{System Architecture of Digital Twin Platform}

\begin{figure}[t]
    \centerline{\includegraphics[scale=0.125]{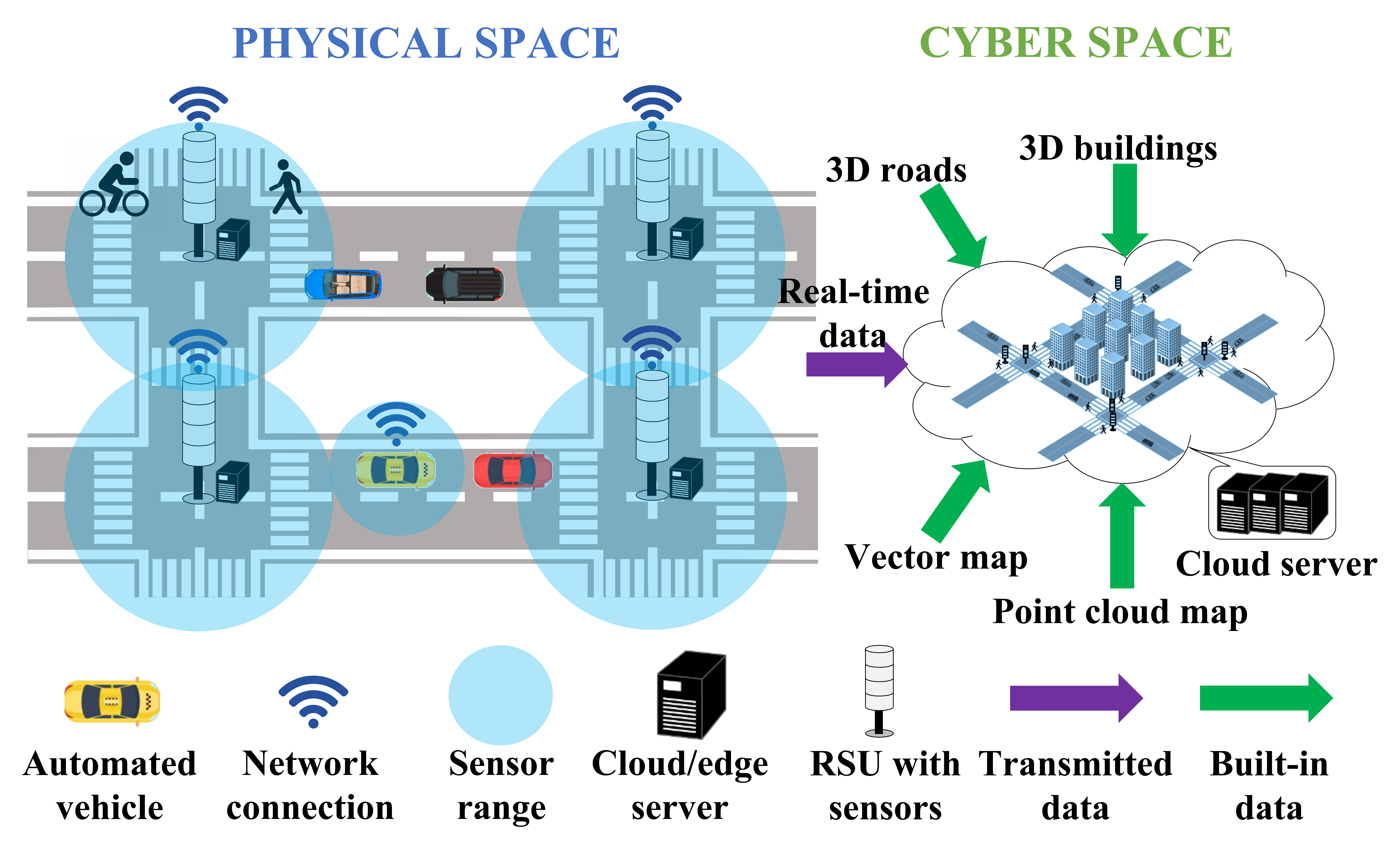}}
    \caption{An overview of the digital twin system model.}
    \label{fig: dt-arc}
\end{figure}

\begin{figure}[t]
\centering
\subfigure[]{
\begin{minipage}[b]{0.22\textwidth}
\includegraphics[width=1\textwidth]{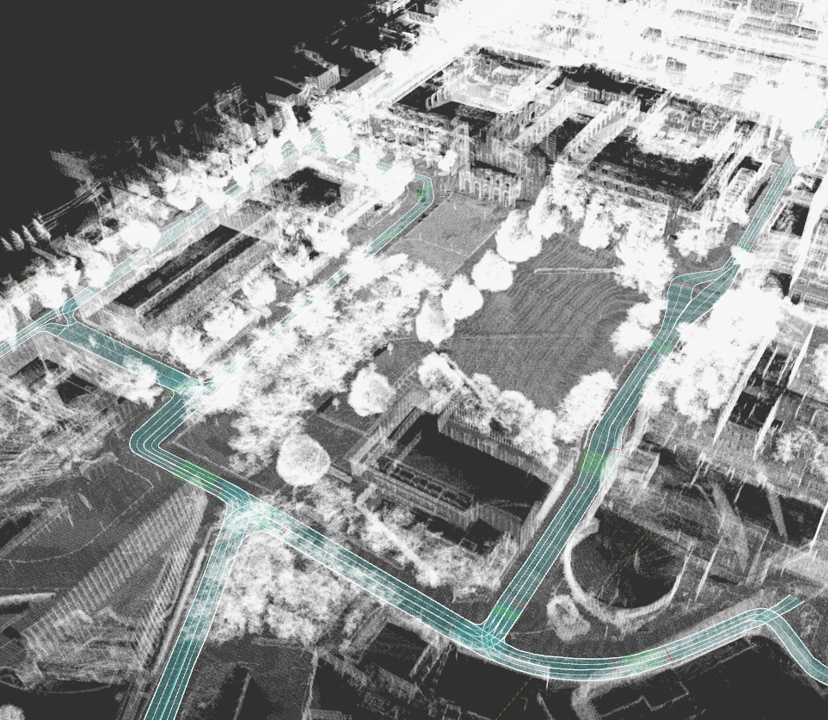} 
\end{minipage}
}
\subfigure[]{
\begin{minipage}[b]{0.22\textwidth}
\includegraphics[width=1\textwidth]{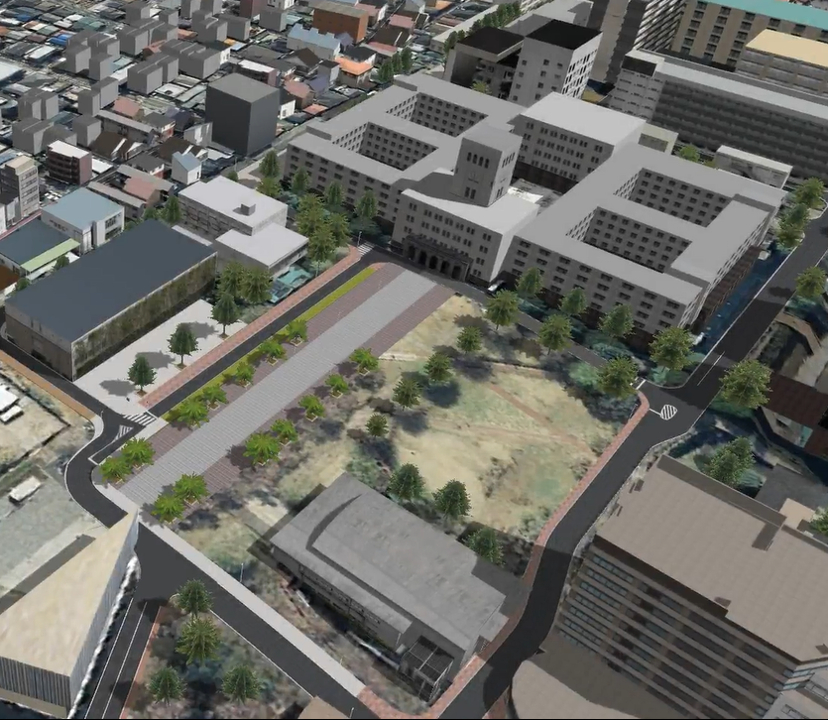} 
\end{minipage}
}
\caption{Built-in data in the digital twin model. (a) Point cloud map and road vector map (b) 3D models of buildings and roads}
\label{fig: built-in}
\end{figure}

In order to make the digital twin platform meet the above requirements, it is important to extract meaningful information from the real traffic and build the digital twin system model. Fig.\ref{fig: dt-arc} gives a conceptual overview of the digital twin system model that consists of a physical space and a cyberspace. The physical space refers to the real world in the sensors’ fields of view, including static (e.g., roads and buildings) and dynamic objects (e.g., vehicles and pedestrians). In cyberspace, some built-in data that reflects the static entities (e.g., terrain, buildings, and roads) is stored in the cloud. Fig. \ref{fig: built-in} shows the point-cloud map, road vector map, and 3D models of Tokyo Tech. Ookayama campus as an example. Regarding the dynamic objects, i.e., the real-time traffic flows in our case, it is necessary to perceive and detect them with the help of sensors and computing devices. 

\begin{figure*}[t]
    \centerline{\includegraphics[scale=0.35]{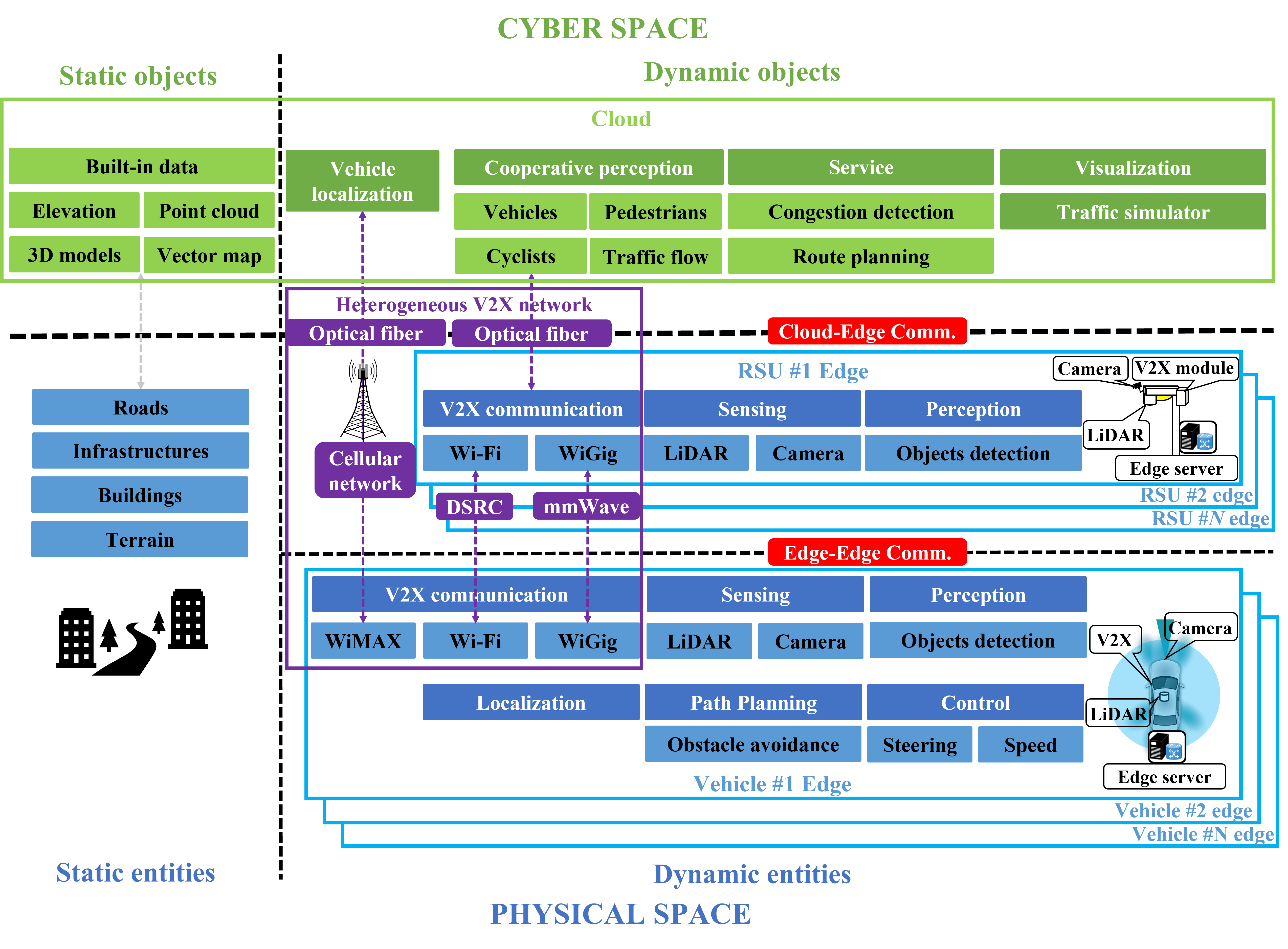}}
    \caption{System architecture of smart mobility platform}
    \label{fig: overall-arc}
\end{figure*}

The overall system architecture is shown in Fig. \ref{fig: overall-arc}. We take advantage of cloud/edge communication and computing capabilities, by distributing different autonomous driving functions and services over the cloud and edge planes. An automated vehicle software system is composed of a series of subsystems, each responsible for a different task. Five modules can form a completely autonomous system, i.e., navigation, perception, localization, planning, and control. In the proposed digital twin platform, some large-scale and computation-intensive tasks that need global information are placed on the cloud (e.g., route planning), while some delay-sensitive tasks with strict real-time requirements are allocated to the edge planes (e.g., localization, motion planning, and motion control). As for the perception module, a larger sensing range implies consuming more computing resources. To ensure a sufficiently large sensing range and sufficient computing resources at the same time, the perception module is allocated to both the cloud and edge servers.

In the RSU edges, we use two types of sensors, i.e., LiDARs and cameras, to obtain some raw data from the physical world. Then we deploy an object detection algorithm to identify and localize dynamic objects, i.e., traffic participants in this study. Compared with the detection results, the amount or size of raw sensor data is dramatically huge, which leads to large latency during the upload processes. Thus, only the processed sensor data will be sent to the cloud to ensure a low delay.

In the vehicle edges, we also applied LiDAR and camera sensors for environment perception. Based on the static point cloud map and real-time LiDAR point cloud, we can localize the vehicle with normal distributions transform (NDT) matching algorithm\cite{b10}. In addition, we use the LiDAR points cloud-based object detection algorithm\cite{b11} to detect surrounding objects, and then the path planning and control modules\cite{b10} will help the vehicle track the planned route and avoid obstacles.

The cloud server, as a global data pool and high-level decision maker, receives processed data uploaded from the edges, synchronizes incoming channels, and locates detected objects based on their relative coordinates to different edge sensors. Then a cooperative perception function can be realized in the cloud plane. Based on the road vector map, we can monitor the real-time traffic condition and traffic flow on each road in the sensing range. Considering both the occupancy level and the distance of each route, a route planner is designed to navigate the vehicle from the origin to the destination.

%%%%%%%%%%%%%%%%%%%%%%%%%%%%%%%%%%%%%%%%%%%%%%%%%%%%%%%%%%%%%%%%%%%%%%%%%%%%%%%%%%%%%%%%%%%%%%%%%%%%%%%%%%%%%%%%%%%%%
Fig.\ref{fig: overall-arc} also shows the heterogeneous V2X communication network for the digital twin platform\cite{b12}. When vehicles pass by RSU-less areas (e.g., rural/mountain areas), they can report their context information such as location, velocity, and direction to the cloud server using cellular networks. Whenever RSUs exist, the vehicles can switch to dedicated short-range communication (DSRC) to broadcast their information, which is then forwarded to the cloud by RSU receivers. The high-level planned route is also distributed by DSRC. Moreover, when the vehicles approach the coverage of millimeter-wave (mmWave) communication, ultra-high-speed and low-latency mmWave V2X can be employed. This heterogeneous V2X network ensures reliable tracking of vehicular contexts in the digital twin platform.    
%%%%%%%%%%%%%%%%%%%%%%%%%%%%%%%%%%%%%%%%%%%%%%%%%%%%%%%%%%%%%%%%%%%%%%%%%%%%%%%%%%%%%%%%%%%%%%%%%%%%%%%%%%%%%%%%%%%%%%

\section{Proof-of-Concept}

In this section, we study the performance of the digital twin platform. A proof-of-concept field test was designed and conducted in a real traffic environment to demonstrate the feasibility of the proposed platform.

\begin{table*}[t]
\centering
\caption{HARDWARE FOR DIGITAL TWIN PLATFORM}
\label{tab:my-table}
\begin{tabular}{c|ccc}
\hline
\multirow{2}{*}{Type}                                                            & \multicolumn{3}{c}{Description}                                                                                                                                    \\ \cline{2-4} 
                                                                                 & Device name                             & Specifications                                                                              & \multicolumn{1}{l}{Amount} \\ \hline
Vehicles                                                                         & RoboCar                                 & Automated vehicle with driving controller                                                   & 1                          \\ \hline
\multirow{2}{*}{Sensors}                                                         & RS-LiDAR-32                             & Position: RSU edges, Range: 200 m,  Accuracy: $\pm$ 3 cm, Rotation speed: 10/20 Hz         & 1                          \\
                                                                                 & RS-LiDAR-80                             & Position: Vehicle edge, Range: 230 m,  Accuracy: $\pm$ 3 cm, Rotation Speed: 5/10/20 Hz    & 4                          \\ \hline
\multirow{3}{*}{Communication}                                                   & WiMAX NIC                               & Position: Vehicle edges, Downlink:120 Mbit/s, Uplink: 60 Mbit/s, Maximum coverage: 30 miles & 2                          \\
                                                                                 & Wi-Fi router                            & Position: RSU \& Vehicle Edges, Tri-Band: 5.72 GHz, 144 ch (Replacement for DSRC)           & 5                          \\
                                                                                 & MmWave antenna                          & Position: RSU \& Vehicle edges, Scan angle:$\pm$ 17 deg (Hor.), $\pm$ 4.5 deg (Ver.)                & 5                          \\ \hline
\multirow{3}{*}{\begin{tabular}[c]{@{}c@{}}Cloud \& Edge\\ Servers\end{tabular}} & Jetson AGX Orin                         & Function: RSU edges, OS: Ubuntu 20.04 (JetPack 5.0.2), ROS: Galactic, Autoware. Universe    & 4                          \\
                                                                                 & Autoware PC                             & Function: Vehicle edge, OS: Ubuntu 16.04, ROS: Kinetic, Autoware. AI                        & 1                          \\
                                                                                 & \multicolumn{1}{l}{Digital twin engine} & Function: Cloud, OS: Ubuntu 18.04, ROS: Melodic, Autoware. AI                               & 1                          \\ \hline
\end{tabular}
\end{table*}

\subsection{Implementation}

\begin{figure}[t]
\centering
\subfigure[]{
\begin{minipage}{0.32\textwidth}
\includegraphics[width=1\textwidth]{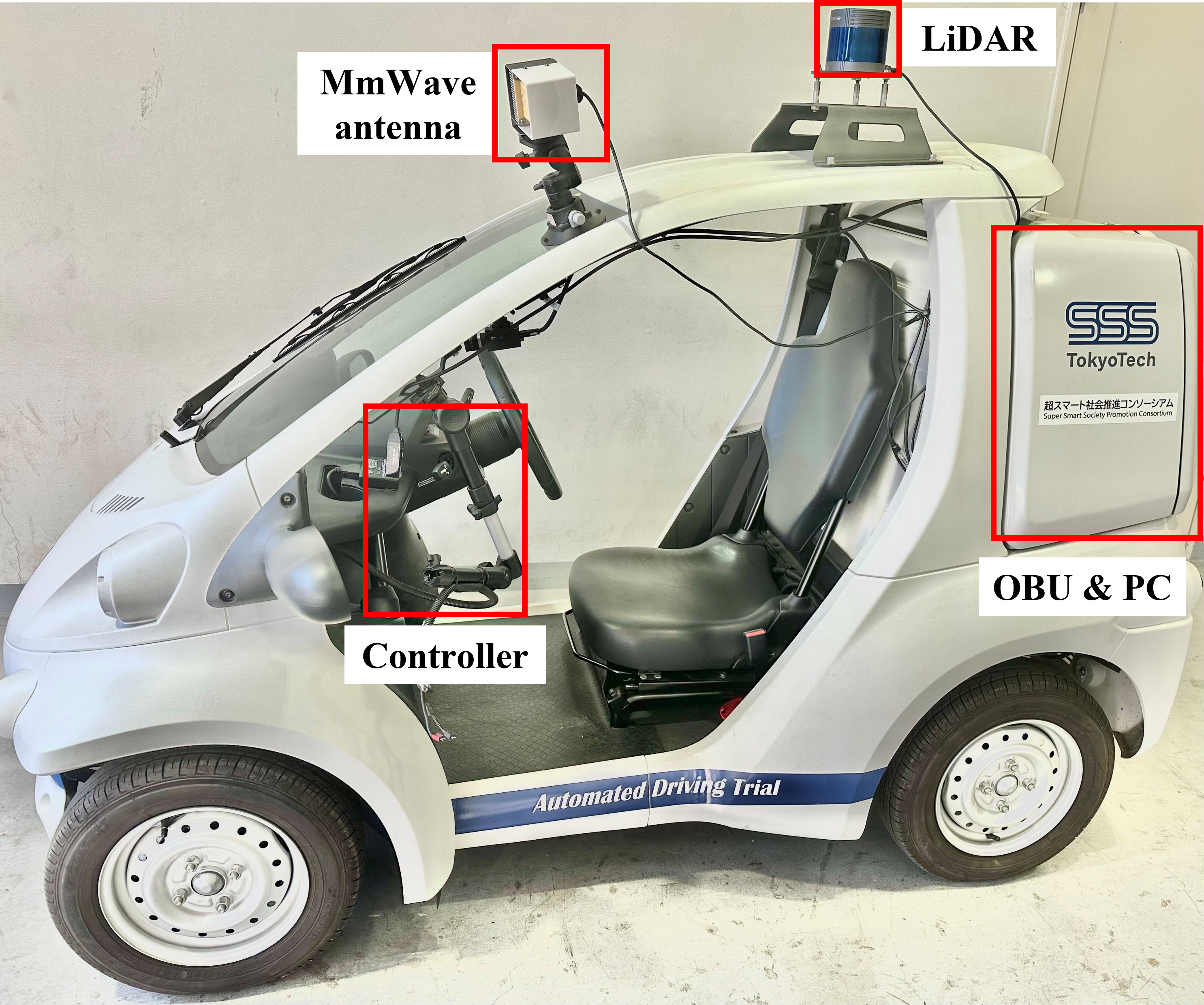} 
\end{minipage}
}
\subfigure[]{
\begin{minipage}{0.12\textwidth}
\includegraphics[width=1\textwidth]{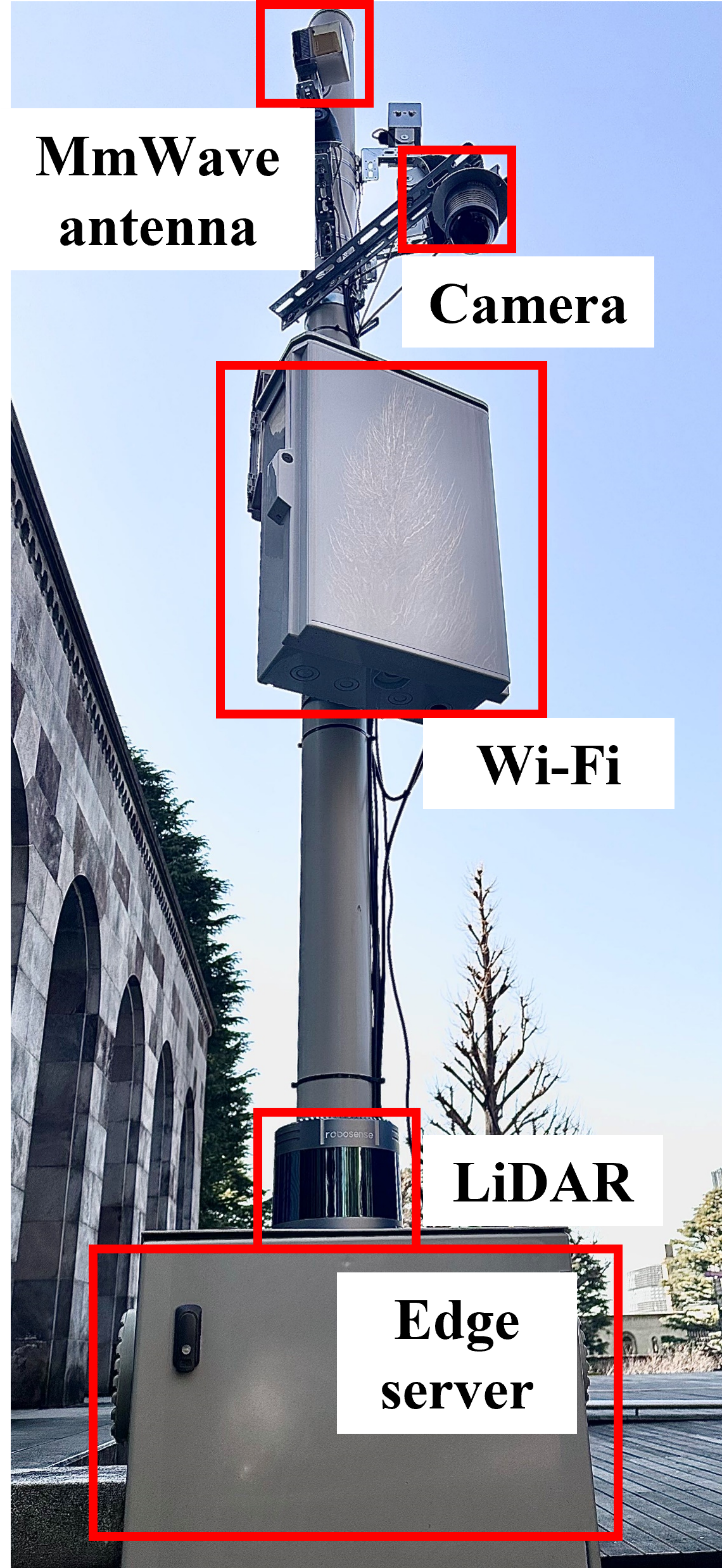} 
\end{minipage}
}
\caption{Experiment devices. (a) Outlook of automated vehicle, (b) Outlook of RSU.}
\label{fig: vehicle}
\end{figure}

The required hardware to be deployed on the edge and cloud is shown in Table I. The automated vehicle, in Fig. \ref{fig: vehicle} (a), is equipped with a LiDAR sensor, mmWave antenna, and Wi-Fi router (which works in 5.72 GHz, 144 ch, as a replacement for DSRC). A PC with Autoware installed is used to drive the automated vehicle by sending control commands to the controller area network (CAN) bus, e.g., the steering angle and the adjustments of acceleration/deceleration pedals, obtaining CAN messages, and processing sensor data. The onboard sensor is 32-layer LiDAR, which is used for vehicle localization and environment perception. We installed four RSU facilities with LiDAR sensors, edge computing devices, and communication equipment as shown in Fig. \ref{fig: vehicle} (b). The sensors used in the experiment on RSUs are 80-layer LiDARs. Compared with the 32-layer LiDAR, the point cloud obtained by the 80-layer LiDAR is denser, so it is more conducive for object detection and recognition. In addition, we use NVIDIA Jetson as an edge computing device for processing raw data and detecting objects.

As for software installation, we fully utilize Autoware \cite{b13} in our system. Autoware enables the research and development of autonomous driving systems in a broad range of applications and consists of all the functionality necessary for automated vehicles, so it is convenient to reorganize the logical structure of automated driving functions for digital twin modeling. More specifically, using sensors and the perception module of Autoware, the traffic environment is perceived, and the traffic participants are detected within the sensor range, respectively for the automated vehicle and RSUs. In addition, Autoware is built on Robot Operating System (ROS), and the nature of ROS makes it easy to deploy in and communicate among distributed computers \cite{b14}. One master machine/computer with bi-directional connectivity between the master and clients enables the running of ROS across multiple machines. In our system, the RSU and vehicle edge computing are regarded as ROS clients, while the central cloud server works as a ROS master. In the cloud server, the results of cooperative perception and detection are visualized in a 3D visualization tool for ROS, i.e., RViz. 

\begin{figure}[t]
    \centerline{\includegraphics[scale=0.5]{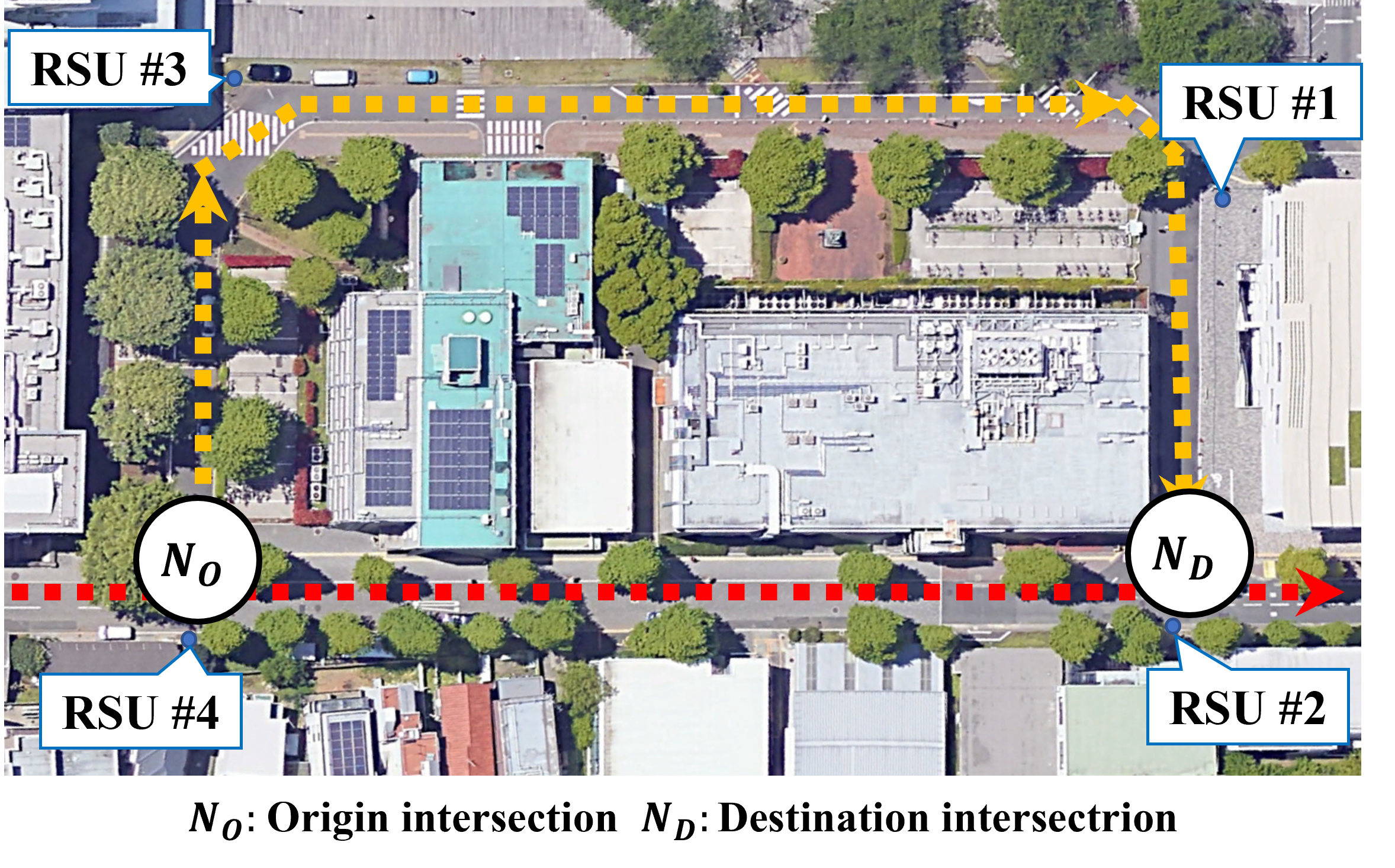}}
    \caption{Overview of the outdoor testing field.}
    \label{fig: field}
\end{figure}

Fig. \ref{fig: field} shows the outdoor field test environment. We selected a road network consisting of two routes and four intersections, and four RSU facilities are located at four intersections. The intersection at the lower-left corner is regarded as an entrance node, while the lower-right intersection is considered an exit node. The red line refers to the shorter default route, while the yellow line shows the alternative route with a longer distance.

\subsection{Test Results and Evaluation}

\begin{figure}[t]
\centering
\subfigure[]{
\begin{minipage}[b]{0.44\textwidth}
\includegraphics[width=1\textwidth]{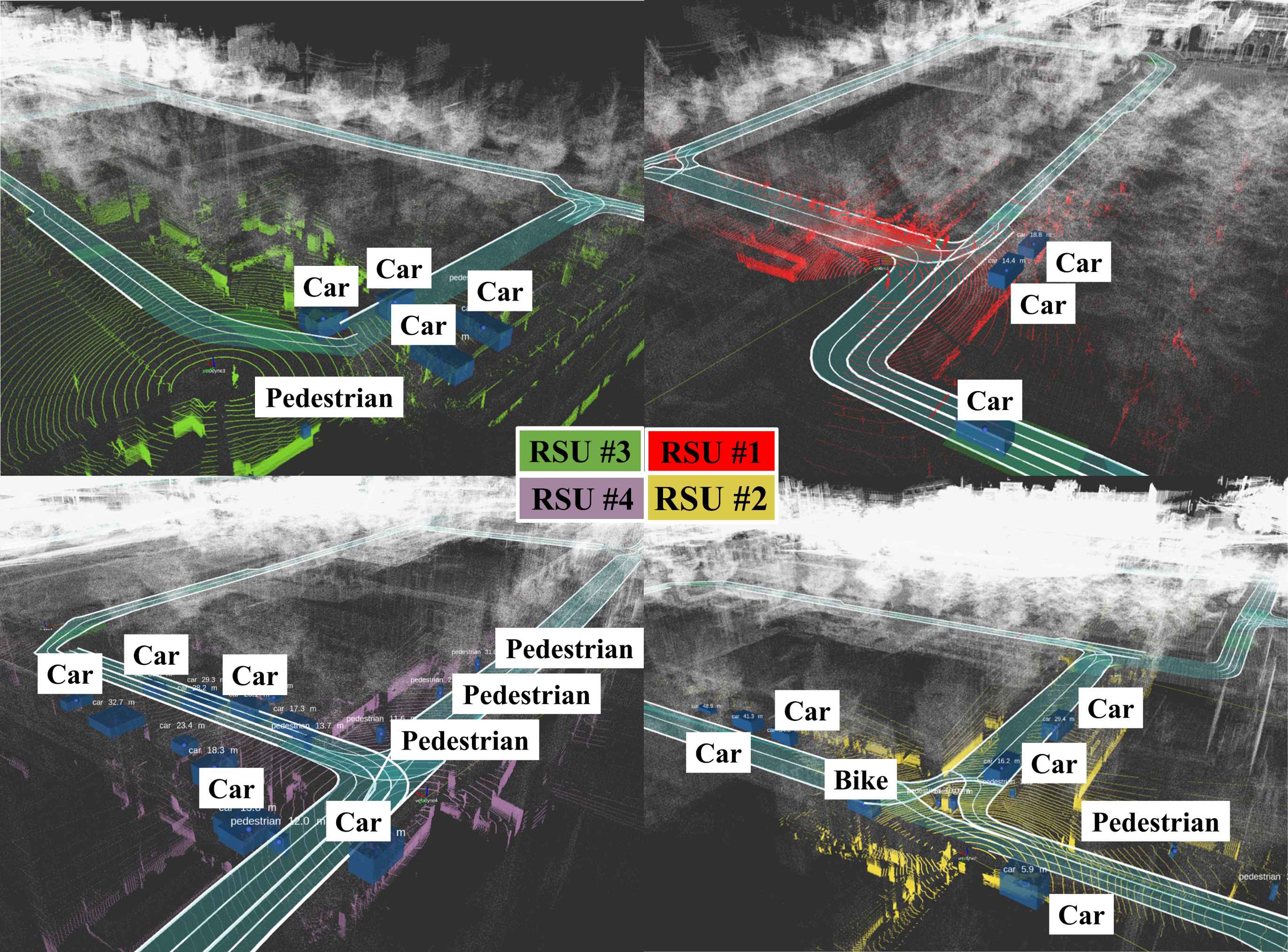} 
\end{minipage}
}
\subfigure[]{
\begin{minipage}[b]{0.44\textwidth}
\includegraphics[width=1\textwidth]{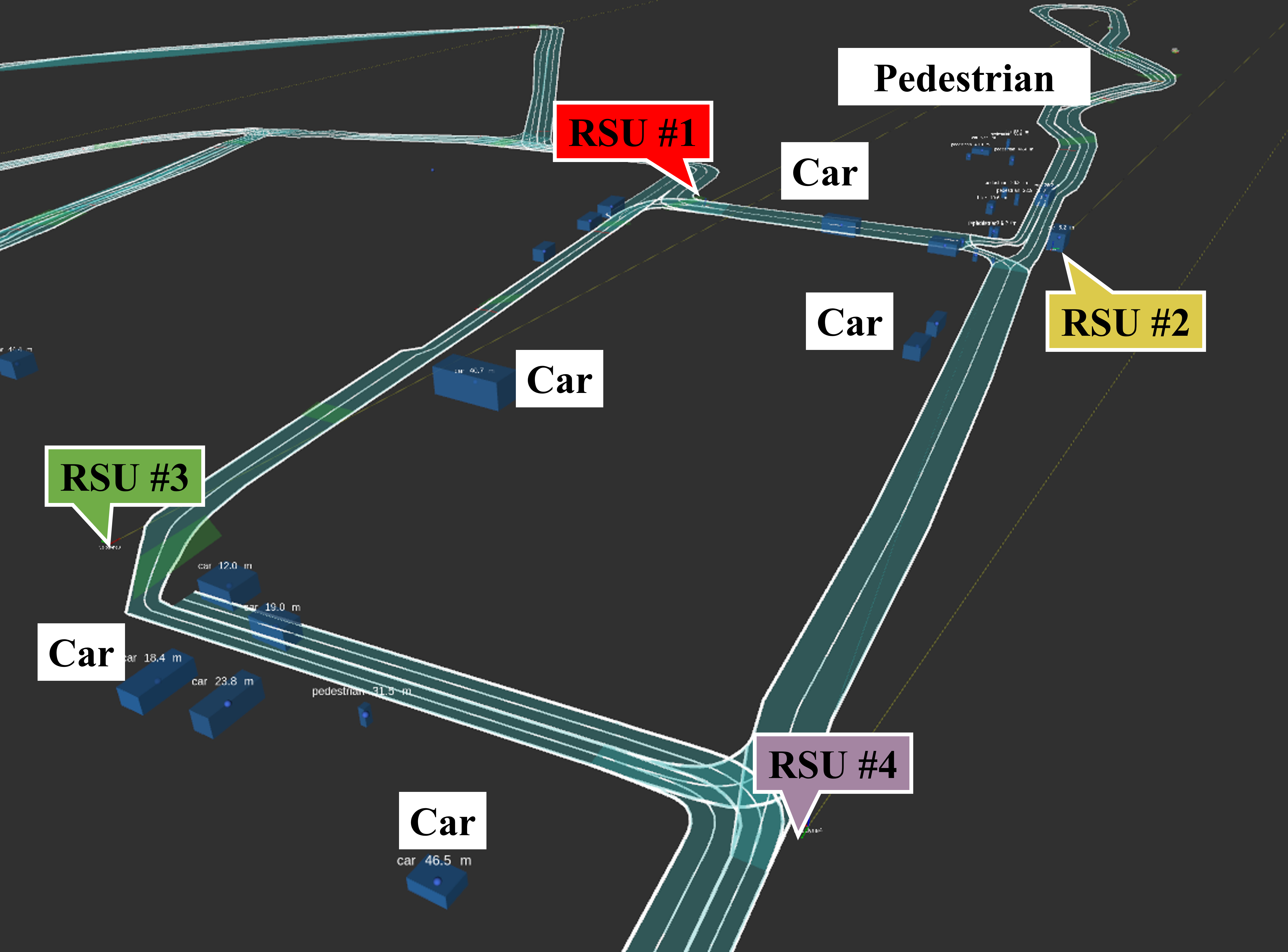} 
\end{minipage}
}
\caption{Environment perception on cloud and RSU edges. (a) Raw data and object detection on RSU edges, (b) Fused object detection on the cloud}
\label{fig: perception}
\end{figure}

An illustration of how the digital twin platform provides route planning services is presented. In phase 1, raw data and the detection results on four RSU edges are shown in Fig. \ref{fig: perception} (a). Some blue cuboids located in the LiDARs range indicate to the detected objects, such as vehicles, pedestrians, and cyclists, as well as their locations and approximate dimensions. In phase 2, after receiving the detection results from RSU edges, the cloud server locates detected objects based on their relative coordinates of the different LiDAR sensors. The blue cuboids in Fig. \ref{fig: perception} (b) indicate the fused detection results. 

\begin{figure}[t]
\centering
\subfigure[]{
\begin{minipage}[b]{0.44\textwidth}
\includegraphics[width=1\textwidth]{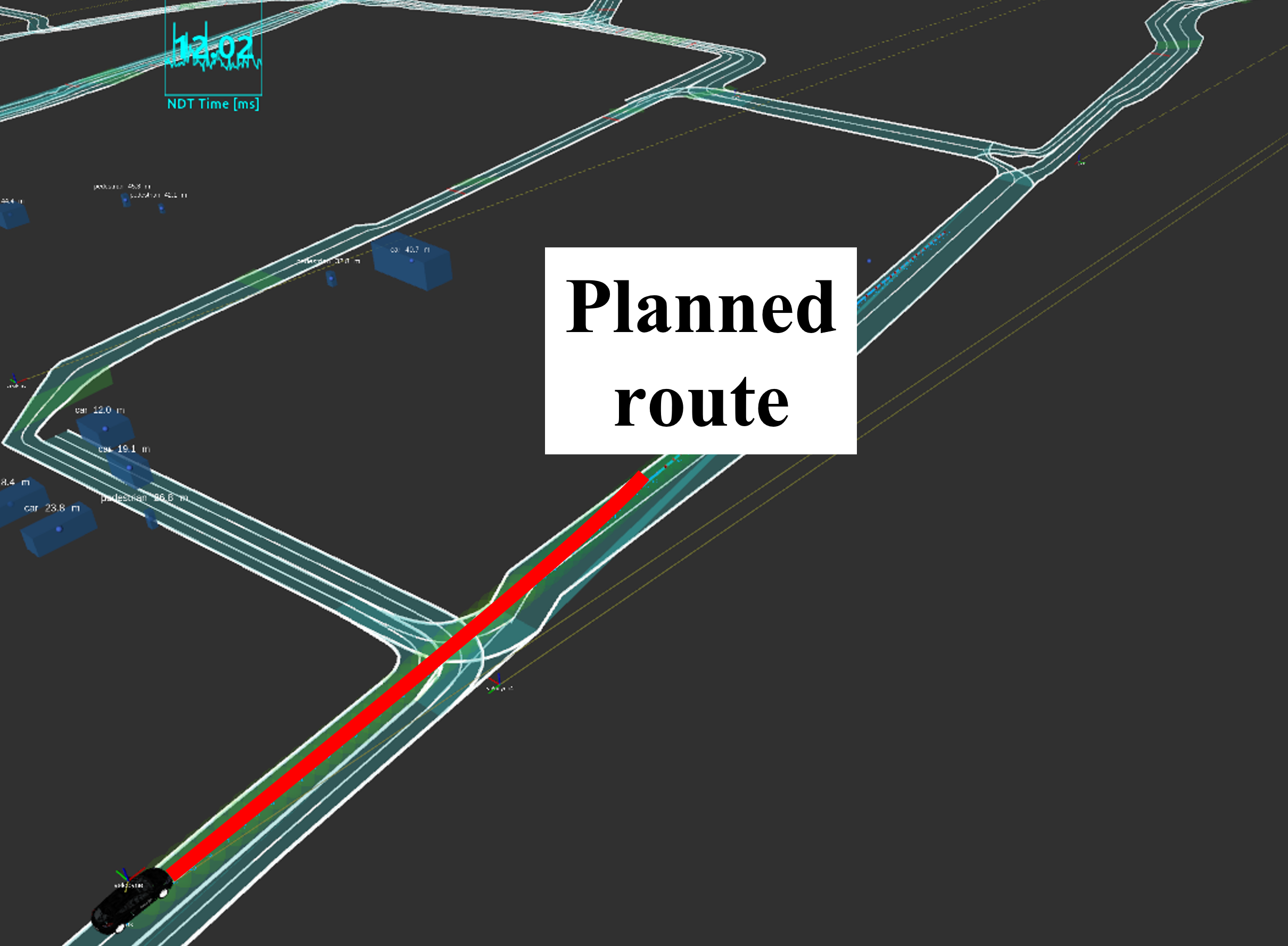} 
\end{minipage}
}
\subfigure[]{
\begin{minipage}[b]{0.44\textwidth}
\includegraphics[width=1\textwidth]{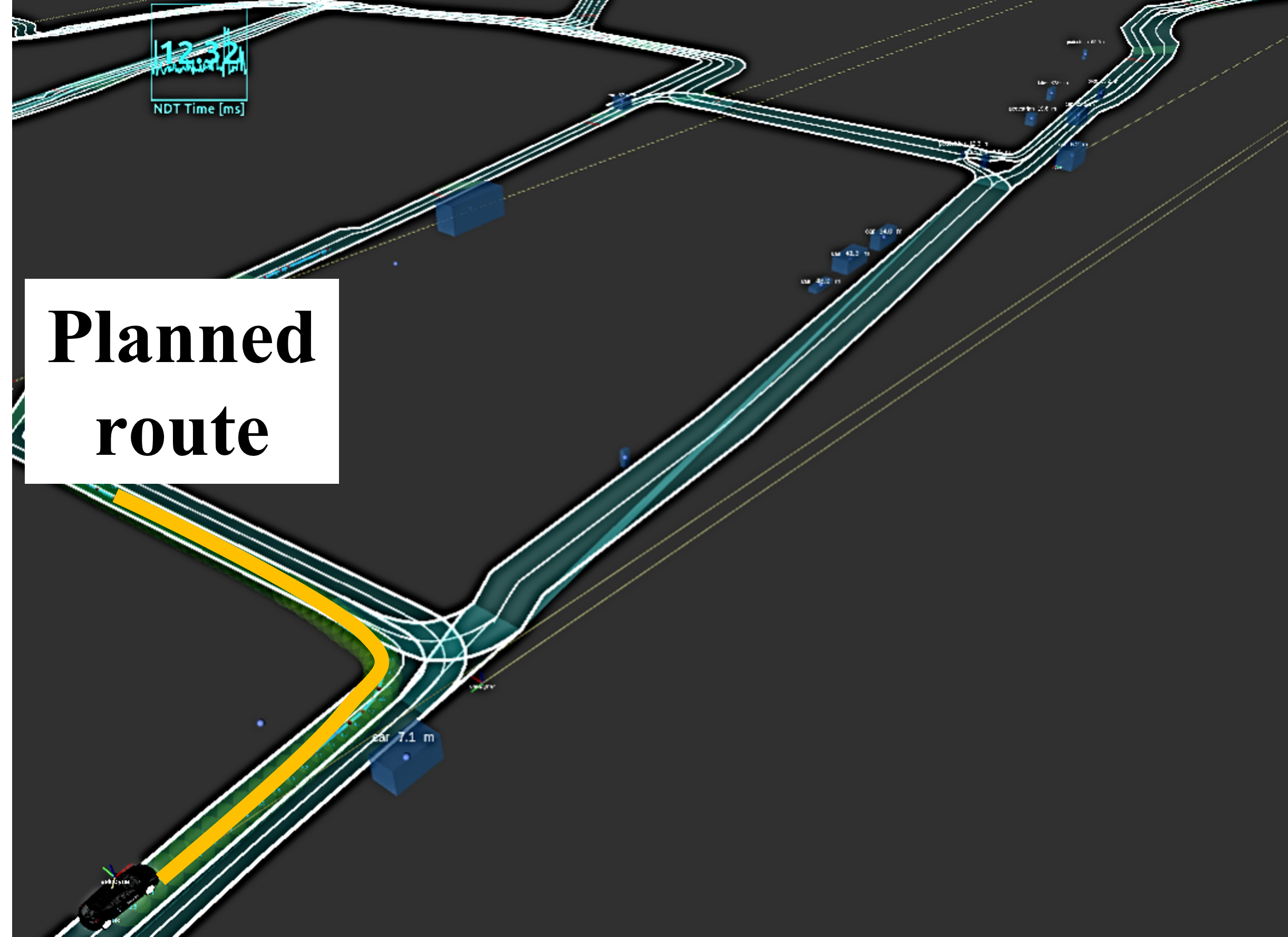} 
\end{minipage}
}
    \caption{Route planning results. (a) Planning with the default route, (b) Planning with the alternative route}
\label{fig: cloud results}
\end{figure}

Considering both the road occupancy level and the distance of each route, the route planning function will choose a route for the vehicle. When there is no excessive traffic flow on the shorter route, the red route in Fig. \ref{fig: cloud results}(a) will be planned to navigate the ego vehicle from the origin towards the destination using the default path. When the occupancy level on the default route is too high, the yellow route in Fig. \ref{fig: cloud results}(b) will be generated to help the ego vehicle to avoid overcrowded traffic on the default route.

As discussed in Sect. II, before the vehicle arrives at the lower-left intersection, the whole process of re-routing service is divided into three phases: 
\begin{itemize}
    \item[1)] Obtain raw sensor data, identify the type, location, and approximate shape of road participants in the sensor range of RSU edges, and upload the detection result to the cloud.
    \item[2)] The cloud server fuses the detection results, then selects the route.
    \item[3)] Sends the selected route to the vehicle edge.
\end{itemize}

The requirement for latency depends on how long the vehicle takes to arrive at the intersection. A 3-second rule is followed in Japan\cite{b15}. Since the maximum speed in our test field is 20 km/h and the length from vehicle CoG to its front edge $l_{r}$ is 0.7 m, $S$ is determined as 17.367 m, and the total latency should be less than 3.126 s. Table II presents the time latency in the proposed digital twin platform, including edge computation, cloud computation, uploading, downloading, and Wi-Fi communication. Edge computing is responsible for the deep learning-based obstacle detection module, which requires an enormous amount of computing resources and leads to large computational time. In the cloud server, after object localization, we need to analyze the position and motion of these objects one by one, e.g., which road section are the objects moving on, are they moving or stationary, and will they affect the speed and trajectory of the ego vehicle, etc. So cloud computing is the most time-consuming part of the platform. The total delay from getting the raw sensor data on RSU edges to receiving the planned route on the vehicle edge is much less than 3.126 s. Therefore, the performance of the proposed digital twin platform meets all the functional and time requirements.

\begin{table}[t]
\centering
\caption{MEASURED TIME LATENCY}
\label{tab:my-table}
\begin{tabular}{c|cccc}
\hline
               & \multicolumn{4}{c}{Latency (ms)} \\ \cline{2-5} 
               & Min    & Avg.   & Max   & Mean dev.    \\ \hline
Edge comp.     & 102    & 107   & 173   & 3.68    \\ \hline
Cloud comp.    & 181    & 188   & 207   & 5.21    \\ \hline
Up \& download & 2.43   & 2.61  & 2.69  & 0.0812  \\ \hline
Wi-Fi comm.    & 1.81   & 15.8  & 105   & 20.3    \\ \hline
Total          & 287   & 243  & 488   & 29.3    \\ \hline
\end{tabular}
\end{table}

\section{Conclusion}

This paper presents a digital twin platform for smart mobility route planning. Utilizing sensors and edge computing on RSUs and connected vehicles, the physical world is projected into the digital world in real time. A smart mobility platform was built to realize cooperative environment perception for traffic monitoring and route planning to help the automated vehicle avoid overcrowded traffic. A proof-of-concept test with a real vehicle in real traffic is conducted to validate the platform from the perspectives of function and delay. Studying the improvements brought to vehicle commuting efficiency and the benefits brought to the whole traffic with the proposed digital twin platform is one of our future works.

\section*{Acknowledgment}

This work was supported by the MEXT Doctoral Program for World-leading Innovative \& Smart Education for Super Smart Society (WISE-SSS), JST SPRING (Grant Number JPMJSP2106), and  NICT-JUNO (Promotion of Commissioned Research for Advanced Communications and Broadcasting Research and Development (\#22404)).

\vspace{12pt}
\end{document}